\documentclass[11pt,a4paper]{article}
\usepackage[hyperref]{emnlp2020}
\usepackage{times}
\usepackage{inconsolata}
\usepackage{latexsym}

\usepackage[ruled,vlined,noend]{algorithm2e}
\usepackage{amsthm}
\theoremstyle{definition}
\newtheorem{definition}{Definition}[]

\usepackage{microtype}

\aclfinalcopy 
\def\aclpaperid{2535} 

\usepackage{amsmath}
\usepackage{mathtools}
\usepackage{nicefrac}
\usepackage{enumitem}
\usepackage{amsfonts}
\usepackage{framed}
\usepackage{xcolor}
\usepackage{xspace}
\usepackage{booktabs}
\usepackage{tikz}
\usepackage{array}
\usepackage{tikz-dependency}
\usetikzlibrary{positioning,arrows}
\setlist[itemize,enumerate]{leftmargin=*}

\DeclareMathOperator{\proj}{\Pi}

\DeclareMathOperator{\conv}{conv}
\DeclareMathOperator*{\argmin}{argmin}
\DeclareMathOperator*{\argmax}{argmax}

\newcommand{\map}{{\tt MAP}}
\newcommand{\marg}{{\tt Marg}}
\newcommand{\smap}{{\tt SparseMAP}}
\newcommand{\ZZ}{\mathcal{Z}}
\newcommand{\MM}{\mathcal{M}}
\newcommand{\pfrac}[2]{\frac{\partial #1}{\partial #2}}
\newcommand\defeq{\coloneqq}
\newcommand\reals{\mathbb{R}}

\newcommand*{\wrt}{\textit{w.\hspace{.07em}r\hspace{.07em}.t.}\@\xspace}
\newcommand*{\eg}{\textit{e.\hspace{.07em}g.}\@\xspace}
\newcommand*{\ie}{\textit{i.\hspace{.07em}e.}\@\xspace}

\definecolor{downstreamcol}{RGB}{192,41,66}
\definecolor{intermcol}{RGB}{83,119,122}

\title{Understanding the Mechanics of SPIGOT:\\Surrogate Gradients for Latent Structure Learning}

\author{Tsvetomila Mihaylova \\
  Instituto de Telecomunica\c{c}\~oes \\
  Instituto Superior Técnico\\
  Lisbon, Portugal \\
  \texttt{tsvetomila.mihaylova@lx.it.pt} \\\And
  Vlad Niculae\textsuperscript{$\dagger$} \\
  Informatics Institute \\
  University of Amsterdam \\
  The Netherlands\\
  \texttt{v.niculae@uva.nl}\And
  Andr\'{e} F.~T. Martins \\
  Instituto de Telecomunica\c{c}\~oes \\
  LUMLIS (Lisbon ELLIS Unit) \\
  Instituto Superior Técnico \& Unbabel \\
  Lisbon, Portugal\\
  \texttt{andre.t.martins@tecnico.ulisboa.com}}

\date{}

\begin{document}
\maketitle
\begingroup
\renewcommand\thefootnote{}%
\footnotetext{\textsuperscript{$\dagger$}Work partially done while VN
was at the Instituto de Telecomunica\c{c}\~oes, Lisbon.}
\endgroup
\begin{abstract}
Latent structure models are a powerful tool for modeling language data:
they can mitigate the error propagation and annotation bottleneck in pipeline
systems, while simultaneously uncovering linguistic insights about the data.
One challenge with end-to-end training of these models is the argmax operation, which has null gradient.
In this paper, we focus on \emph{surrogate gradients}, a popular strategy to deal with this problem. We explore latent structure learning through the angle of \emph{pulling back}
the downstream learning objective. In this paradigm, we discover
a principled motivation for both the straight-through estimator (STE) as well as
the recently-proposed SPIGOT---a variant of STE for structured models.
Our perspective leads to
new algorithms in the same family.
We empirically compare the known and the novel pulled-back estimators against
the popular alternatives, yielding new insight for practitioners
and revealing intriguing failure cases.
\end{abstract}

\section{Introduction}

Natural language data is \emph{rich in structure}, but most of the structure is
not visible at the surface.  Machine learning models tackling high-level language
tasks would benefit from uncovering underlying structures such as trees,
sequence tags, or segmentations.  Traditionally, practitioners turn to
\emph{pipeline} approaches where an external, pretrained model is used to
predict, \eg, syntactic structure. 
The benefit of this approach is that the predicted tree is readily available for inspection, but the downside is that the errors 
can easily propagate throughout the pipeline and require further attention \citep{finkel2006solving,sutton2005joint,toutanova2005effective}.
In contrast, deep neural architectures
tend to eschew such preprocessing, and instead learn soft hidden
representations, not easily amenable to visualization and analysis.

The best of both worlds would be to \emph{model structure as a latent variable},
combining the transparency of the pipeline approach with the end-to-end
unsupervised representation learning that makes deep models appealing.
Moreover, large-capacity model tend to rediscover structure from scratch
\citep{tenney-etal-2019-bert}, so structured latent variables
may reduce the required capacity.

Learning with discrete, combinatorial latent variables is, however, challenging,
due to the intersection of
\emph{large cardinality}
and
\emph{null gradient} issues.
For example, when learning a latent dependency tree,
the latent parser must choose among an exponentially large set of possible
trees; what's more, the parser may only learn from gradient information from the
downstream task. If the highest-scoring tree is selected using an \emph{argmax}
operation, the gradients will be zero, preventing learning.

One strategy for dealing with the null gradient issue is to use a
\textbf{surrogate gradient}, explicitly overriding the zero gradient
from the chain rule, as if a different computation had been performed.
The most commonly known example is the \emph{straight-through estimator}
\citep[STE;][]{bengio2013estimating},
which pretends that the \emph{argmax} node was instead an
\emph{identity} operator.
Such methods lead to a fundamental mismatch between the objective and the learning algorithm.
The effect of this mismatch  is still insufficiently understood,
and the design of successful new variants is therefore challenging.
For example, the recently-proposed SPIGOT method \citep{spigot}
found it beneficial to use a projection as part of the surrogate gradient.

In this paper, we study surrogate gradient methods for deterministic learning with
discrete structured latent variables. Our contributions are:

\begin{itemize}
\item We propose a novel motivation for surrogate gradient methods, based on optimizing
a \textbf{pulled-back loss}, thereby inducing pseudo-supervision on the latent variable.
This leads to new insight into both STE and SPIGOT.
\item We show how our framework may be used to derive new surrogate gradient methods,
by varying the loss function or the inner optimization algorithm used for inducing the pseudo-supervision.
\item We experimentally validate our discoveries on a controllable experiment as well as on English-language
sentiment analysis and natural language inference, comparing against stochastic and relaxed alternatives,
yielding new insights, and identifying noteworthy failure cases.
\end{itemize}

While the discrete methods do not outperform the relaxed alternatives using the same building
\linebreak
blocks, we hope that our interpretation and insights would trigger future
latent structure research.

The code for the paper is available on \url{https://github.com/deep-spin/understanding-spigot}.

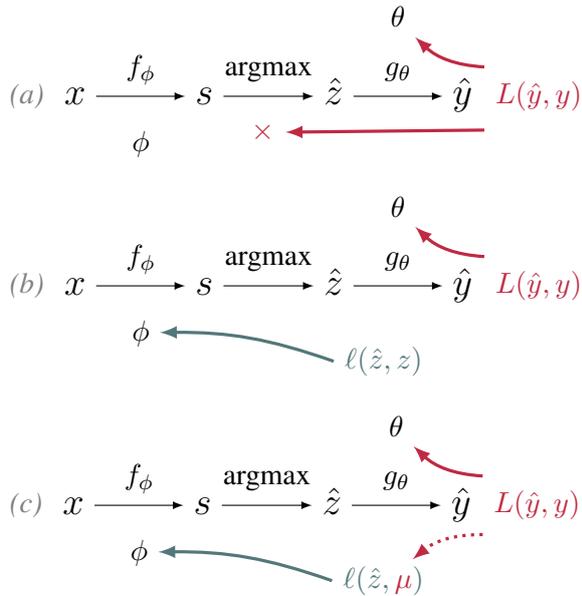
\begin{figure}[t]
\newcommand{\scaffold}{%
\node[nd] at (0, 0) (x) {$\strut x$};
\node[nd] at (1, 0) (s) {$\strut s$};
\node[nd] at (2, 0) (z) {$\strut \hat{z}$};
\node[nd] at (3, 0) (y) {$\strut \hat{y}$};

\node[downstream,right of=y] (dloss) {$\strut L(\hat{y},y)$};

\draw[->] (x) -- node[midway,above] (f) {$\strut f_\phi$} (s);
\draw[->] (s) -- node[midway,above] (am) {\strut argmax} (z) ;
\draw[->] (z) -- node[midway,above] (g) {$\strut g_\theta$}(y);

\node[above=1pt of g] (theta) {$\theta$};
\node[below of=f] (phi) {$\phi$};

\draw[->,downstream] (dloss.north west) to[bend left=20] (theta);
}

\tikzset{%
    >=latex,
    nd/.style={font=\Large},
    downstream/.style={downstreamcol,very thick},
    interm/.style={intermcol, very thick}
}

\begin{tikzpicture}[scale=1.7,baseline]
\scaffold
\draw[downstream,->] ([yshift=-1pt]dloss.south west) to
([yshift=-1pt,xshift=-15pt]z.south east) node[anchor=east] {$\times$};

\node[left=0pt of x,color=gray] {\emph{(a)}};
\end{tikzpicture}

\vspace{.5\baselineskip}

\begin{tikzpicture}[scale=1.7,baseline]
\scaffold
\node[interm,below of=z,anchor=west] (iloss) {$\ell(\hat{z}, z)$};
\draw[->,interm] (iloss.west) to[bend right=10] (phi);
\node[left=0pt of x,color=gray] {\emph{(b)}};
\end{tikzpicture}

\vspace{.5\baselineskip}

\begin{tikzpicture}[scale=1.7,baseline]
\scaffold
\node[interm,below of=z,anchor=west] (iloss) %
  {$\ell(\hat{z}, \textcolor{downstreamcol}{\mu})$};
\draw[->,interm] (iloss.west) to[bend right=10] (phi);
\node[left=0pt of x,color=gray] {\emph{(c)}};
\draw[->,downstream,dotted] (dloss.south west) to[bend right=20] %
  ([xshift=-5pt,yshift=-1pt]iloss.north east);
\end{tikzpicture}
\caption{
A model with a discrete latent variable $z$.
Given an input $x$, we assign a score $s_z = [f(x)]_z$ to each choice, and pick
the highest scoring one, $\hat{z}$, to predict $\hat{y} = g_\theta(\hat{z})$.
For simplicity, here $g_\theta$ does not access $x$ directly.
(a). Since argmax has null gradients,
the encoder parameters $\phi$ do not receive updates.
(b). If ground truth supervision were available for the latent $z$, $\phi$
could be trained jointly with an auxiliary loss.
(c). As such supervision is not available, we induce a best-guess label
$\mu$ by pulling back the downstream loss.
This strategy recovers the STE and SPIGOT estimators.}
\label{fig:pullback-loss}
\end{figure}

\section{Related Work}

Discrete latent variable learning is often tackled in
\textbf{stochastic computation graphs}, by
estimating the gradient of an expected loss.
An established method is the score function estimator (SFE)
\citep{Glynn:1990:LRG:84537.84552,williams1992simple,kleijnen1996optimization}.
SFE is widely used in NLP, for tasks including
minimum risk training in NMT \cite{shen-etal-2016-minimum,wu-etal-2018-study}
and latent linguistic structure learning
\citep{latentparse,havrylov-etal-2019-cooperative}.
In this paper, we focus on the alternative strategy of
\textbf{surrogate gradients}, which allows learning
in deterministic graphs with  discrete, argmax-like nodes, rather than
in stochastic graphs.
Examples are the \textbf{straight-through estimator (STE)} \citep{hinton2012ste,bengio2013estimating} and the
structured projection of intermediate
gradients optimization technique (SPIGOT; \citealt{spigot}).
Recent work focuses on studying and explaining STE. \citet{yin2019understanding} obtained a convergence result in shallow networks for the unstructured case. \citet{cheng2018straight} show that STE can be interpreted as the simulation of the projected Wasserstein gradient flow. STE has also been studied in binary neural networks \citep{hubara2016binarized} and in other applications \citep{tjandra2019end}. Other methods based on the surrogate gradients have been recently explored \citep{poganvcic2019differentiation,meng2020training}.

A popular alternative is to \textbf{relax} an argmax into a continuous transform such
as softmax or sparsemax \citep{sparsemax}, as seen for instance in soft attention mechanisms \citep{transf}, or structured attention networks
\citep{rush,dani,lapata,mensch2018differentiable,sparsemap}.
In-between surrogate gradients and relaxation is \textbf{Gumbel softmax},
which uses the Gumbel-max reparametrization to sample from
a categorical distribution, applying softmax either to relax the mapping or to
induce surrogate gradients \citep{jang2017categorical,maddison2016concrete}. Gumbel-softmax has been successfully applied to latent linguistic structure as well \citep{choi2018learning,maillard2018}. For sampling from a structured variable is required, the \textbf{Perturb-and-MAP} technique \citep{papandreou2011perturb} has been successfully applied to sampling latent structures in NLP applications \citep{Corro2019,Corro2019-acl}.

\section{Structured Prediction Preliminaries}

We assume a general latent structure model involving input variables $x \in
\mathcal{X}$, output variables $y \in \mathcal{Y}$, and latent discrete
variables $z \in \mathcal{Z}$.  We assume that $\mathcal{Z} \subseteq
\{0,1\}^K$, where $K \le |\mathcal{Z}|$ (typically, $K \ll |\mathcal{Z}|$):
\ie, the latent discrete variable $z$ can be represented as a $K$-th
dimensional binary vector. This often results from a decomposition of a
structure into parts: for example, $z$ could be a dependency tree for a sentence
of $L$ words, represented as a vector of size $K=O(L^2)$, indexed by pairs of
word indices $(i,j)$, with $z_{ij} = 1$ if arc $i \rightarrow j$ belongs to the
tree, and $0$ otherwise.
This allows us to define the \textbf{score} of a structure as the sum of the
scores of its parts. Given a vector
$s \in \reals^K$,
containing scores for all possible parts, we define
\begin{equation}
\operatorname{score}(z) \defeq s^\top z.
\end{equation}

\paragraph{Notation.}
We denote by $e_k$ the one-hot vector with all zeros except in the $k$\textsuperscript{th} coordinate.
We denote the
simplex by $\triangle^{|\mathcal{Z}|}
\defeq \{p \in \mathbb{R}^{|\mathcal{Z}|}\!\mid\!p \ge 0, \sum_{z \in \mathcal{Z}} p(z) = 1\}.$
Given a distribution $p \in \triangle^{|\mathcal{Z}|}$, the expectation of a function
$h:\mathcal{Z}\rightarrow \mathbb{R}^D$ under $p$
is $\mathbb{E}_{z \sim p}[h(z)] \defeq \sum_{z \in \mathcal{Z}} p(z) h(z)$.
We denote the convex hull of the (finite) set $\mathcal{Z} \subseteq \mathbb{R}^K$
by $\mathrm{conv}(\mathcal{Z}) \defeq \left\{\mathbb{E}_{z \sim p}[z] \mid
p \in \triangle^{|\mathcal{Z}|}\right\}$.
The euclidean projection of $s$ onto a set $\mathcal{D}$ is
$\proj_{\mathcal{D}}(s) \defeq \argmin_{d \in \mathcal{D}} \|s - d\|_2$.

\paragraph{Background.}
In the context of structured prediction,
the set $\MM \defeq \conv(\ZZ)$ is known as
the \emph{marginal polytope}, since any point inside it can be
interpreted as some marginal distribution over parts of the structure (arcs)
under some distribution over structures.
There are three relevant problems that may be formulated in a structured
setting:
\begin{itemize}
\item Maximization (MAP inference): finds a highest scoring structure,
$\map(s) \coloneqq
\displaystyle \argmax_{z \in \ZZ} s^\top z.$

\item Marginal inference: finds the (unique) marginals induced by
the scores $s$,
corresponding to the Gibbs distribution where $p(z) \propto \exp\big(%
\operatorname{score}(z)\big)$.
The solution maximizes the entropy-regularized objective
\begin{equation}
\marg(s)\coloneqq\argmax_{\mu \in \MM} s^\top\mu + \tilde{H}(\mu),
\end{equation}
where $\tilde{H}$ is the maximum entropy among all distributions over
\emph{structures} that achieve marginals $\mu$
\citep{Wainwright2008}:
\begin{equation}
\tilde{H}(\mu) \coloneqq
\max_{\substack{p \in \triangle^{|\ZZ|} \\ \mathbb{E}_p[z] = \mu}}
-\sum_{z \in \ZZ} p(z) \log p(z).
\end{equation}

\item SparseMAP: finds the (unique) \emph{sparse} marginals induced by
the scores $s$, given by a Euclidean projection onto $\MM$:
\citep{sparsemap}
\end{itemize}
\begin{equation}\label{eq:sparsemap}
\begin{aligned}
\smap(s) &\coloneqq \proj_{\MM}(s) \\
&=\argmax_{\mu \in \MM} s^\top\mu - \frac{1}{2} \|\mu\|^2.
\end{aligned}
\end{equation}

\paragraph{Unstructured setting.}
As a check, we consider the encoding of a categorical variable with $K$
distinct choices, encoding each choice as a one-hot vector $e_k$ and setting
$\ZZ = \{ e_1, \dots, e_K\}$. In this case, $\conv(\ZZ) = \triangle^K.$
The optimization problems above then recover some well known transformations, as
described in Table~\ref{table:structured_vs_unstructured}.
\begin{table}[ht]
\small
\centering
\begin{tabular}{@{}r l l@{}}
& unstructured & structured \\
\hline
vertices & $e_k$ & $z_k$ \\
interior points & $p$ & $\mu$ \\
maximization & \tt argmax & \tt MAP \\
expectation & {\tt softmax} & \marg \\
Euclidean projection & \tt sparsemax & \tt SparseMAP \\
\hline
\end{tabular}
\caption{Building blocks for latent structure models.}
\label{table:structured_vs_unstructured}
\end{table}

\section{Latent Structure Models}\label{section:latent_structure_model}

Throughout,
we assume a classifier parametrized by $\phi$ and $\theta$, which consists of three parts:
\begin{itemize}
    \item An {\bf encoder function} $f_{\phi}$ which, given an input  $x \in \mathcal{X}$, outputs a vector of ``scores'' $s \in \mathbb{R}^K$, as $s = f_{\phi}(x)$;
    \item An {\bf argmax node} which, given these scores, outputs the highest-scoring structure:
    \begin{equation}\label{eq:argmax}
    \hat{z}(s) \coloneqq \argmax_{z \in \mathcal{Z}} s^\top z = \map(s)\,;
    \end{equation}
    \item A {\bf decoder function} $g_{\theta}$ which, given $x \in \mathcal{X}$ and $z \in \mathcal{Z}$, makes a prediction $\hat{y} \in \mathcal{Y}$ as $\hat{y} = g_{\theta}(x, z)$. We will sometimes write $\hat{y}(z)$ to emphasize the dependency on $z$.
    For reasons that will be clear in the sequel, we
must
assume that the decoder also accepts
\emph{average structures},
\ie, it can also output predictions
$g_{\theta}(x, \mu)$ where $\mu \in \mathrm{conv}(\mathcal{Z})$
is a convex combination (weighted average) of structures.
\end{itemize}
Thus, given input $x \in \mathcal{X}$, this network predicts:
\begin{equation}
    \hat{y} = g_\theta\left(x, \overbrace{\argmax_{z \in \mathcal{Z}} f_\phi(x)^\top z}^{\hat{z}(s)}\right).
\end{equation}
To train this network, we minimize a loss function  $L(\hat{y}, y)$, where
$y$ denotes the target label; a common example is the negative log-likelihood loss.

The gradient \wrt the decoder parameters, $\nabla_\theta L(\hat{y}, y)$,
is easy to compute using automatic differentiation on $g_\theta$.
The main challenge is to propagate gradient information \textbf{through the argmax
node} into the encoder parameters. Indeed,
\begin{equation*}
    \nabla_\phi L(\hat{y}, y) = \frac{\partial f_\phi(x)}{\partial \phi} \underbrace{\frac{\partial \hat{z}(s)}{\partial s}}_{=0} \nabla_z L(\hat{y}(\hat{z}), y) = 0,
\end{equation*}
so no gradient will flow to the encoder.
We list below the three main categories of approaches that tackle this issue.

\paragraph{Introducing stochasticity.}
Replace the argmax node by a stochastic node where $z$ is modeled as a random
variable $Z$ parametrized by $s$ (\eg, using a Gibbs distribution).
Then, instead of optimizing a deterministic loss $L(\hat{y}(\hat{z}), y)$,
optimize the
\textbf{expectation} of the loss under the predicted distribution:
\begin{equation}\label{eqn:exploss}\mathbb{E}_{Z \sim p(z; s)}[L(\hat{y}(Z), y)].\end{equation}
The expectation ensures that the gradients are no longer null.
This is sometimes referred to as \emph{minimum risk training} \citep{smith2006minimum,stoyanov2011empirical},
and typically optimized using
the \emph{score function estimator}
\citep[SFE;][]{Glynn:1990:LRG:84537.84552,williams1992simple,kleijnen1996optimization}.

\paragraph{Relaxing the argmax.}
Keep the network deterministic, but relax the argmax node
into a continuous function, for example replacing it with softmax or sparsemax
\citep{sparsemax}. In the structured case, this gives rise to structured
attention networks \citep{rush} and their SparseMAP variant
\citep{sparsemap}.
This corresponds to moving the expectation inside the loss,
optimizing
$L\big(\hat{y}(\underbrace{\mathbb{E}_{Z \sim p(z; s)}[Z]}_{\mu}), y\big)$.
\paragraph{Inventing a surrogate gradient.}
Keep the argmax node and perform the usual forward computation,
but backpropagate a different, non-null gradient in the backward pass.
This is the approach underlying straight-through estimators
\citep{hinton2012ste, bengio2013estimating} and SPIGOT \citep{spigot}.
This method introduces a mismatch between the measured objective and the
optimization algorithm. In this work, we proposed a novel, principled
justification for inducing surrogate gradients.
In what follows, we assume that:
\begin{itemize}
    \item We can compute the gradient
\begin{equation}\label{eq:gamma}
\gamma(\mu) := \nabla_\mu L(\hat{y}(\mu), y)\,,
\end{equation}
for any $\mu$, usually by automatic differentiation;%
\footnote{This gradient would not exist if the decoder $g_\theta$ were
defined only at vertices $z\in\ZZ$ and not mean points $\mu \in \MM$.}
    \item We want to replace the null gradient $\nabla_s L(\hat{y}(\hat{z}),
y)$ by a surrogate $\tilde{\nabla}_s L(\hat{y}(\hat{z}), y)$.
\end{itemize}

\section{SPIGOT as the Approximate Optimization of a Pulled Back Loss}

We next provide a novel interpretation of SPIGOT as the minimization of
a ``pulled back'' loss.
SPIGOT uses the surrogate gradient:
\begin{equation}\label{eq:spigot}
\begin{aligned}
\tilde{\nabla}_s L(\hat{y}(\hat{z}), y) &=
\hat{z} - \proj_{\MM} \left(\hat{z} - \eta \gamma\right)\\
&= \hat{z} - \texttt{SparseMAP}(\hat{z} - \eta \gamma),
\end{aligned}
\end{equation}
highlighting that SparseMAP \citep{sparsemap} computes an
Euclidean projection (Eq.~\ref{eq:sparsemap}).

\subsection{Intermediate Latent Loss}

To begin, consider a much simpler scenario:
if we had supervision for the latent variable $z$
(\eg, if the true label $z$ was revealed to us), we could define an
{\bf intermediate loss} $\ell(\hat{z}, z)$
which would induce nonzero updates to the encoder parameters.
Of course, we do not have access to this $z$. Instead, we consider the following alternative:

\begin{framed}
\theoremstyle{definition}
\begin{definition}[Pulled-back label]
A guess $\mu \in \MM =\conv(\ZZ)$ for what the unknown $z \in \ZZ$ should be, informed by the downstream loss.
\end{definition}
\end{framed}
\noindent Figure~\ref{fig:pullback-loss} provides the intuition of the pulled-back label and loss.
We take a moment to justify picking $\mu \in \MM$ rather than directly in $\ZZ$.
In fact, if $K=|\ZZ|$ is small, we can enumerate all possible values of $z$ and
define the guess as the latent value minimizing the downstream loss, $\mu =
\argmin_{z \in \ZZ} L(\hat{y}(z), y)$.
This is sensible, but intractable in the structured case. Moreover, early on in the training process, while $g_\theta$ is untrained, the maximizing vertex carries little information.
Thus, for robustness and tractability, we
allow for some uncertainty by picking
a convex combination $\mu \in \MM$ so as to approximately minimize
\begin{equation}\label{eq:pullback_loss}
    \mu \approx \argmin_{\mu \in \MM} L(\hat{y}(\mu), y).
\end{equation}
For most interesting predictive models $\hat{y}(\mu)$ (\eg, deep networks),
this optimization problem is non-convex and lacks a closed form solution.
One common strategy is the {\bf projected gradient algorithm}
\citep{goldstein1964convex,levitin1966constrained}, which, in addition to
gradient descent, has one more step: projection of the updated point on the constraint set. It iteratively performs the following updates:
\begin{equation}\label{eq:projected_gradient}
\mu^{(t+1)} = \proj_{\MM} \left(\mu^{(t)} - \eta_t \gamma(\mu^{(t)})\right)\,,
\end{equation}
where $\eta_t$ is a step size and $\gamma$ is as in Eq.~\ref{eq:gamma}.
With a suitable choice of step sizes, the projected gradient algorithm converges
to a local optimum of Eq.~\ref{eq:pullback_loss} \citep[Proposition
2.3.2]{bertsekas-nonlin}.
In the sequel, for simplicity we use constant $\eta$.
If we initialize $\mu^{(0)} = \hat{z} = \argmax_{z \in \mathcal{Z}} s^\top z$,
{\bf a single iteration} of projected gradient yields the
guess:
\begin{equation}\label{eq:single_projected_gradient}
    \mu^{(1)} = \proj_{\MM} \big(\hat{z} - \eta \gamma(\hat{z})
\big)\,.
\end{equation}
Treating the induced $\mu$ as if it were the ``ground truth'' label of $z$,
we
may train the encoder $f_\phi(x)$ by {\bf supervised learning}.
With a {\bf perceptron loss},
\begin{eqnarray}\label{eq:perceptron_loss}
    \ell_{\mathrm{Perc}}(\hat{z}(s), \mu) &=&
    \max_{z \in \mathcal{Z}} s^\top z - s^\top \mu\nonumber\\
    &=& s^\top \hat{z} - s^\top \mu\,,
\end{eqnarray}
a single iteration yields the gradient:
\begin{equation}\label{eq:perc_grad}
\nabla_s \ell_{\mathrm{Perc}}(\hat{z}, \mu^{(1)}) = \hat{z} -
\mu^{(1)}
\,,
\end{equation}
which is precisely the SPIGOT gradient surrogate in Eq.~\ref{eq:spigot}.
This leads to the following insight into how SPIGOT updates the encoder parameters:
\begin{framed}%
\noindent SPIGOT
minimizes the {\bf perceptron loss} between $z$ and a
pulled back target computed by {\bf one projected gradient step}
on $\displaystyle\min_{\mu \in \MM} L(\hat{y}(\mu), y)$ starting at
$\hat{z}=\map(s)$.
\end{framed}
\noindent This construction suggests possible alternatives, the first of which
uncovers a well-known algorithm.

\begin{description}[style=unboxed,leftmargin=0cm]
    \item[Relaxing the {\boldmath $\MM$} constraint.]
    The constraints in Eq.~\ref{eq:pullback_loss} make the optimization problem more
    complicated. We relax them and define
    $\mu \approx \argmin_{\mu \in \textcolor{purple}{\mathbb{R}^K}}
L(\hat{y}(\mu), y)$.
    This problem still requires iteration, but the projection step can now
    be avoided. One iteration of gradient descent yields $\mu^{(1)} = \hat{z} -
    \eta \gamma$. The perceptron update then recovers
a novel derivation of
\textbf{straight-through} with identity (STE-I),
where the backward pass acts as if $\pfrac{\hat{z}(s)}{s} \stackrel{!}{=}
\mathrm{Id}$ \citep{bengio2013estimating},
\begin{equation}\label{eq:ste}
\nabla_s \ell_{\mathrm{Perc}}(\hat{z}, \mu^{(1)}) = \hat{z} - (
\hat{z} - \eta \gamma) = \eta \gamma.
\end{equation}
This leads to the following insight into straight-through and its relationship
to SPIGOT:
\begin{framed}%
\noindent Straight-through (STE-I) minimizes the {\bf perceptron loss} between $z$ and a pulled back
target computed by {\bf one gradient step} on $\displaystyle \min_{\mu \in
\mathbb{R}^K} L(\hat{y}(\mu), y)$ starting at $\hat{z}=\map(s)$.
\end{framed}
\end{description}
From this intuition, we readily obtain new surrogate gradient methods, which we
explore below.

\section{New Surrogate Gradient Methods}

\begin{description}[style=unboxed,leftmargin=0cm]
\item[Multiple gradient updates.] Instead of a single projected
gradient step, we could run multiple steps of Eq.~\ref{eq:projected_gradient}. We would expect this to yield
a better approximation of $\mu$.
This comes at a computational cost:
each update involves running a forward and backward pass in the decoder $g_\theta$ with the current guess
$\mu^{(t)}$, to obtain $\gamma(\mu^{(t)}) \defeq \nabla_\mu L\big(\hat{y}(\mu^{(t)}), y\big)$.
\item[Different initialization.] The projected gradient update in
Eq.~\ref{eq:single_projected_gradient} uses $\mu^{(0)} = \hat{z} = \argmax_{z
\in \mathcal{Z}} s^\top z$ as the initial point. This is a sensible choice, if
we believe the encoder prediction $\hat{z}$ is close enough to the
optimal $\mu$, and it is computationally convenient, because
the forward pass uses $\hat{z}$, so
$\gamma(\hat{z})$ is readily available in the backward pass,
thus the first inner iteration comes for free.
However, other initializations are possible, for example $\mu^{(0)} =
\marg(s)$ or $\mu^{(0)} = 0$, at the cost of an extra computation of $\gamma(\mu^{(0)})$.
In this work, we do not consider alternate initializations for their own sake; they are
needed for the following two directions.
\item[Different intermediate loss: SPIGOT-CE.] For simplicity, consider the
unstructured case where $\MM = \triangle$, and use the initial guess
${\mu}^{(0)} = \texttt{softmax}(s)$.
Replacing $\ell_\text{Perc}$ by the cross-entropy loss
$\ell_{\mathrm{CE}}({\mu}^{(0)}, \mu^{(1)}) = -\sum_{k=1}^K \mu_k \log {\mu}^{(0)}_k$
yields
\begin{equation}
\nabla_s \ell_{\text{CE}}({\mu}^{(0)}, \mu^{(1)}) = {\mu}^{(0)} - \proj_{\triangle}
( \mu^{(0)} - \eta\gamma ).
\end{equation}
In the structured case, the corresponding loss is the CRF loss \citep{lafferty2001conditional}, which
corresponds to the KL divergence between two distributions over structures.
In this case, we initialize $\mu^{(0)} = \marg(s)$ and update
\begin{equation}\label{eq:spigotce}
\nabla_s \ell_{\text{CE}}(\mu^{(0)}, \mu^{(1)}) = {\mu}^{(0)} - \proj_{\MM}({\mu}^{(0)}- \eta \gamma).
\end{equation}
    \item[Exponentiated gradient updates: SPIGOT-EG.]
In the unstructured case, optimization over $\MM = \triangle$ can also be tackled via
the exponentiated gradient (EG) algorithm \citep{kivinen1997exponentiated},
which minimizes
Eq.~\ref{eq:pullback_loss} with the following multiplicative update:
\begin{equation}\label{eq:exponentiated_gradient}
    \mu^{(t+1)} \propto \mu^{(t)}\odot \exp(-\eta_t \nabla_\mu L(\hat{y}(\mu^{(t)}), y)),
\end{equation}
where $\odot$ is elementwise multiplication and thus each iterate $\mu^{(t)}$ is
strictly positive, and normalized to be inside $\triangle$.
EG cannot be initialized on the boundary of $\triangle$,
so again we must take $\mu^{(0)} = \texttt{softmax}(s)$.
A single iteration of EG yields:
\begin{eqnarray}\label{eq:exponentiated_gradient_single_update}
    \mu^{(1)} &\propto& {\mu}^{(0)} \odot\exp(-\eta\gamma)\nonumber\\
    &=& \texttt{softmax}(\log {\mu}^{(0)} -\eta \gamma)\nonumber\\
    &=& \texttt{softmax}(s -\eta \gamma).
\end{eqnarray}
It is natural to use the cross-entropy loss, giving
\begin{equation}
\nabla_s \ell_\text{CE}({\mu}^{(0)}\!\!,\mu^{(1)})\!=\!{\mu}^{(0)} - \texttt{softmax}(s - \eta\gamma),
\end{equation}
\ie, the surrogate gradient is the difference between the softmax prediction
and a ``perturbed'' softmax.
To generalize to the structured case, we observe that
both EG and projected gradient are instances of mirror descent under
KL divergences \citep{teboullemd}. Unlike the unstructured case, we must
iteratively keep track of both \emph{perturbed scores} and \emph{marginals},
since $\marg^{-1}$ is non-trivial.
This leads to the following mirror descent algorithm:
\begin{equation}\label{eq:mirror}
\begin{aligned}
s^{(0)} &= s,~~~\mu^{(0)} = \marg(s^{(0)})\,, \\
s^{(t+1)} &= s^{(t)} - \eta \gamma(\mu^{(t)})\,, \\
\mu^{(t+1)} &= \marg(s^{(t)})\,.\\
\end{aligned}
\end{equation}
With a single iteration and the CRF loss, we get
\begin{equation}
\nabla_s \ell_\text{CE} =
\marg(s) - \marg(s - \eta\gamma)\,.
\end{equation}
\end{description}
Algorithm~\ref{algo:forward_backward} sketches the implementation of the proposed surrogate gradients for the structured case.
The forward pass is the same for all variants: given the scores $s$ for the parts
of the structure, it calculates the $\map$ structure $z$. The surrogate gradients are implemented as custom backward passes.
The function \texttt{GradLoss} uses automatic differentiation to compute
$\gamma(\mu)$ at the current guess $\mu$; each call involves thus a forward and
backward pass through $g_{\theta}$.
Due to convenient initialization, the first iteration of STE-I and SPIGOT come
for free, since both $\mu^{(0)}$ and $\gamma(\mu^{(0)})$ are available as a byproduct when computing the
forward and, respectively, backward pass through $g_{\theta}$ 
in order to update $\theta$.
For SPIGOT-CE and SPIGOT-EG, even with $k=1$ we need a second call to the
decoder, since $\mu^{(0)} \neq \hat{z}$,
so an additional decoder call is necessary for obtaining the gradient of the loss with respect to $\mu^{(0)}$.
The unstructured case is essentially identical, with $\marg$
replaced by $\texttt{softmax}$.

\begin{algorithm}[t]
\small
\SetAlgoLined
\SetKwInput{KwInput}{Parameters}
\SetKwFunction{FGradLoss}{GradLoss}
\SetKwFunction{FForward}{Forward}
\SetKwFunction{FBackwardSTE}{BackwardSTE-I}
\SetKwFunction{FBackwardSPIGOT}{BackwardSPIGOT}
\SetKwFunction{FBackwardSPIGOTEG}{BackwardSPIGOT-EG}
\SetKwFunction{FBackwardSPIGOTCE}{BackwardSPIGOT-CE}
\def\algspace{.5\baselineskip}
\SetKwProg{Fn}{Function}{:}{}
\KwInput{step size $\eta$, n.\ iterations $k$}
\vspace{\algspace}
\Fn{\FForward{$s$, $x$, $y$}}{
    \KwRet{%
    $\hat{z} \leftarrow \map(s)$\hfill%
    \tcp*[h]{Eq.\,\eqref{eq:argmax}}%
    }
}
\vspace{\algspace}
\Fn{\FGradLoss{$\mu, x, y$}}{
    \KwRet{$\gamma \leftarrow \nabla_{\mu}L(\hat{y}(\mu), y)$\hfill%
    \tcp*[h]{Eq.\,\eqref{eq:gamma}}%
    }
}
\vspace{\algspace}
\Fn{\FBackwardSPIGOT{$s, x, y$}}{
    ${\mu}^{(0)} = \map(s)$ \\

    \For{$t\gets 1$ \KwTo $k$}{
        $\gamma \leftarrow$ \FGradLoss{${\mu}^{(t-1)}, x, y$} \\
        ${\mu}^{(t)} \leftarrow \Pi_{\MM}({\mu}^{(t-1)} - \eta\gamma)$\hfill%
        \tcp*[h]{Eq.\,\eqref{eq:projected_gradient}}%
    }
    \KwRet{$\mu^{(0)} - {\mu}^{(k)}$\hfill%
    \tcp*[h]{Eq.\,\eqref{eq:perc_grad}}%
    }
}
\vspace{\algspace}
\Fn{\FBackwardSTE{$s, x, y$}}{
    ${\mu}^{(0)} = \map(s)$ 
\hfill\tcp*[h]{Eq.\,\eqref{eq:ste}}\\

    \For{$t\gets1$ \KwTo $k$}{
        $\gamma \leftarrow$ \FGradLoss{${\mu}^{(t-1)}, x, y$} \\
        ${\mu}^{(t)} \leftarrow {\mu}^{(t-1)} - \eta\gamma$
    }
    \KwRet{$\mu^{(0)} - {\mu}^{(k)}$}%
}
\vspace{\algspace}
\Fn{\FBackwardSPIGOTCE{$s, x, y$}}{
    ${\mu}^{(0)} \leftarrow \marg(s)$ 
\hfill\tcp*[h]{Eq.\,\eqref{eq:spigotce}}\\

    \For{$t\gets1$ \KwTo $k$}{
        $\gamma\leftarrow$ \FGradLoss{${\mu}^{(t-1)}, x, y$} \\
        ${\mu}^{(t)} \leftarrow \Pi_{\MM}({\mu}^{(t-1)} - \eta\gamma)$
    }
    \KwRet{${\mu}^{(0)} - {\mu}^{(k)}$}
}
\vspace{\algspace}
\Fn{\FBackwardSPIGOTEG{$s, x, y$}}{
    $({s}^{(0)}, {\mu}^{(0)}) \leftarrow (s, \marg(s))$ 
\hfill\tcp*[h]{Eq.\,\eqref{eq:mirror}}\\

    \For{$t\gets1$ \KwTo $k$}{
        $\gamma\leftarrow$ \FGradLoss{${\mu}^{(t-1)}, x, y$} \\
        $s^{(t)} \leftarrow s^{(t-1)} - \eta\gamma$ \\
        ${\mu}^{(t)} \leftarrow \marg(s^{(t)})$
    }
    \KwRet{${\mu}^{(0)} - {\mu}^{(k)}$}
}
\caption{Surrogate gradients pseudocode:
common forward pass, specialized backward passes.
\label{algo:forward_backward}}
\end{algorithm}

\begin{table*}[ht!]
\centering
\small
\begin{tabular}{ l
r@{\color{gray}\small$\pm$}>{\color{gray}\small}l r@{$\color{gray}\small\pm$}>{\color{gray}\small}l@{\qquad}r@{$\color{gray}\small\pm$}>{\color{gray}\small}l r@{$\color{gray}\small\pm$}>{\color{gray}\small}l }
\toprule & \multicolumn{4}{c}{\textbf{(3 clusters)}} & \multicolumn{4}{c}{\textbf{(10 clusters)}} \\ \textbf{Model} &
\multicolumn{2}{c}{\textbf{Accuracy}} &
\multicolumn{2}{c}{\textbf{V-measure}}&
\multicolumn{2}{c}{\textbf{Accuracy}} &
\multicolumn{2}{c}{\textbf{V-measure}}\\
\midrule
\multicolumn{5}{l}{\emph{Baselines}} \\
Linear model              &    68.05 & 0.09 &    0.00 &  0.00&    60.00 & 0.06 &    0.00 & 0.00 \\
Gold cluster labels       &    92.40 & 0.06 &  100.00 &  0.00&    88.50 & 0.10 &  100.00 & 0.00 \\
\addlinespace[1ex]
\multicolumn{5}{l}{\emph{Relaxed}} \\
Softmax                   &    93.15 & 0.33 &   66.88 &  0.97&    86.45 & 0.33 &   75.07 & 1.18 \\
Sparsemax                 &    92.95 & 0.38 &   71.35 & 16.60&    83.75 & 1.32 &   76.13 & 3.89 \\
*Gumbel-Softmax           &    94.25 & 3.42 &  100.00 &  6.80&    80.45 & 0.77 &   89.68 & 1.10 \\
\addlinespace[1ex]
\multicolumn{5}{l}{\emph{Argmax}} \\
*ST-Gumbel                &    93.85 & 3.25 &  100.00 &  6.80&    81.25 & 0.68 &   91.52 & 1.46 \\
*SFE                      &    68.45 & 0.33 &   47.73 & 17.65&    59.80 & 0.58 &   55.56 & 3.30 \\
*SFE w/ baseline          &    94.20 & 0.08 &  100.00 &  0.00&    84.70 & 0.97 &   96.83 & 0.85 \\
STE-S                     &    86.95 & 4.01 &   84.44 & 11.61&    75.95 & 1.10 &   82.83 & 2.75 \\
STE-I                     &    92.60 & 0.23 &  100.00 &  0.00&    84.50 & 1.43 &   94.48 & 1.35 \\
SPIGOT                    &    77.90 & 1.26 &   20.53 &  1.85&    68.80 & 1.02 &   29.24 & 2.24 \\
SPIGOT-CE                 &    93.40 & 2.64 &   97.08 & 13.92&    83.50 & 0.87 &   94.88 & 1.39 \\
SPIGOT-EG                 &    92.70 & 3.04 &  100.00 &  8.27&    79.40 & 2.03 &   82.29 & 2.15 \\
\bottomrule
\end{tabular}
\caption{Discrete latent variable learning on synthetic data: downstream accuracy and clustering V-measure. Median and standard error reported over four runs.
We mark stochastic methods with *.
}
\label{table:results_unstructured}
\end{table*}

\section{Experiments}

Armed with a selection of surrogate gradient methods, we now proceed to an experimental comparison.
For maximum control, we first study a synthetic unstructured experiment with known data generating process.
This allows us to closely compare the various methods, and to identify basic failure cases.
We then study the structured case of latent dependency trees for sentiment analysis and natural language inference in English.
Full training details are described in Appendix~\ref{appendix:training_details}.

 \begin{figure*}[t]
     \centering
    \centering \footnotesize
    \includegraphics[width=.9\textwidth]{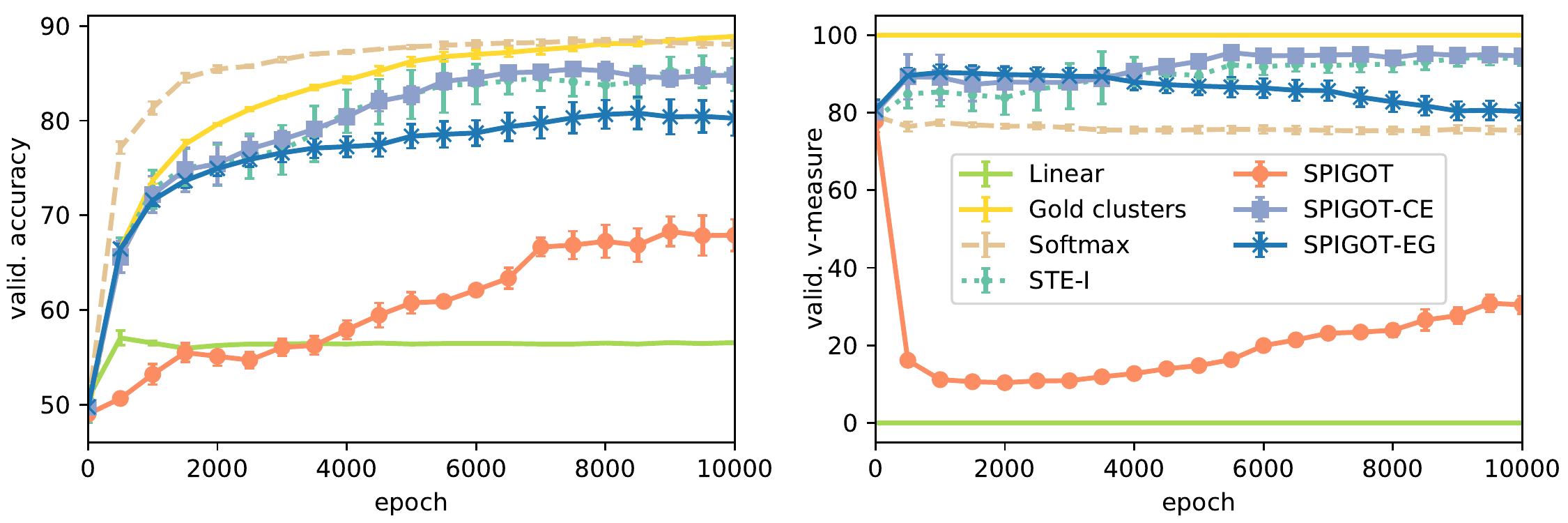}
    \caption{Learning curves on synthetic data with 10 clusters.
Softmax learns the downstream task fast, but mixes the clusters,
yielding poor V-measure. SPIGOT fails on both metrics; STE-I and the novel
SPIGOT-CE work well.
 \label{fig:lcurve}}
 \end{figure*}
\subsection{Categorical Latent Variables}
\label{section:unstructured}

For the unstructured case, we
design a synthetic dataset from
a mixture model
$z \sim \operatorname{Categorical}(\nicefrac{1}{K})$,
$x \sim \operatorname{Normal}(m_z, \sigma I)$,
$y = \operatorname{sign}(w_z^\top x + b_z)$,
where $m_z$ are randomly
placed cluster centers,
and $w_z, b_z$ are parameters of a different ground truth linear model for each cluster.
Given cluster labels,
one could
learn the optimal linear classifier separating the data in that cluster. Without knowing the cluster,
a global linear model cannot fit the data well.
This setup provides a test bed for discrete variable learning, since
accurate clustering leads to a good fit.
The architecture, following
\S\ref{section:latent_structure_model}, is:

\begin{itemize}
    \item \textbf{Encoder}: A linear mapping from the input to a $K$-dimensional score vector:
    $s = f_{\phi}(x) = W_{f}x + b_{f}$, where $\phi = (W_f,b_f) \in \mathbb{R}^{K\times \mathrm{dim}(\mathcal{X})} \times \mathbb{R}^K$ are parameters.
    \item \textbf{Latent mapping}:
$\hat{z} = \rho(s),$
where $\rho$ is $\operatorname{argmax}$ or a continuous relaxation
such as $\texttt{softmax}$ or $\texttt{sparsemax}$.
\item \textbf{Decoder}: A bilinear transformation,
combining the input $x$ and the latent variable $z$:
\[\hat{y} = g_{\theta}(x,\hat{z}) = \hat{z}^\top W_{g} x + b_{g},\]
where $\theta = (W_g,b_g) \in \mathbb{R}^{K\times \mathrm{dim}(\mathcal{X})} \times \mathbb{R}$ are model parameters.
If $\hat{z} = e_k$, this \emph{selects} the $k$\textsuperscript{th} linear model from the rows of $W_{g}$.
\end{itemize}

\noindent We evaluate two baselines: a linear model, and an \emph{oracle} where $g_\theta(x, z)$
has access to the true $z$.
In addition to the methods discussed in the previous section,
we evaluate \emph{softmax} and \emph{sparsemax} end-to-end differentiable relaxations,
and the STE-S variant which uses the \texttt{softmax} backward pass while doing \texttt{argmax} in the forward pass.
We also compare \emph{stochastic} methods, including
score function estimators (with an optional moving average control variate),
and the two Gumbel estimator variants
\citep{jang2017categorical,maddison2016concrete}:
Gumbel-Softmax with relaxed
\texttt{softmax} in the forward
pass, and the other using \texttt{argmax} in the style of STE (hence dubbed ST-Gumbel).

\paragraph{Results.} We compare the discussed methods in Table~ \ref{table:results_unstructured}.
Knowledge of the data-generating process allows us to measure
not only downstream accuracy, but also
\textbf{clustering quality}, by comparing the model predictions with the known true $z$.
We measure the latter via the V-measure \cite{vmeasure},
a clustering score independent of the cluster labels, \ie, invariant to permuting the labels (between 0 and 100, with 100 representing perfect cluster recovery).
The linear and gold cluster oracle baselines confirm that cluster separation is needed
for good performance.
Stochastic models perform well across both criteria.
Crucially, SFE requires variance reduction to performs well, but
even a simple control variate will do.

Deterministic models may be preferable when likelihood assessment or sampling is not
tractable. Among these, {STE-I} and {SPIGOT-\{CE,EG\}} are indistinguishable from the best models.
Surprisingly, the vanilla SPIGOT fails, especially in cluster recovery.
Finally, the
relaxed deterministic models perform
very well on accuracy and learn very fast (Figure~\ref{fig:lcurve}), but appear to rely on mixing clusters, therefore
they remarkably fail to recover cluster assignments.\footnote{With relaxed methods, the V-measure is always calculated using the
argmax, even though $g_\theta$ sees a continuous relaxation.}
This is in line with the structured results of \citet{Corro2019-acl}.
Therefore, if latent structure recovery is less important than downstream accuracy,
relaxations
seem preferable.

\begin{figure}[t]
    \centering \footnotesize
    \includegraphics[width=.49\textwidth]{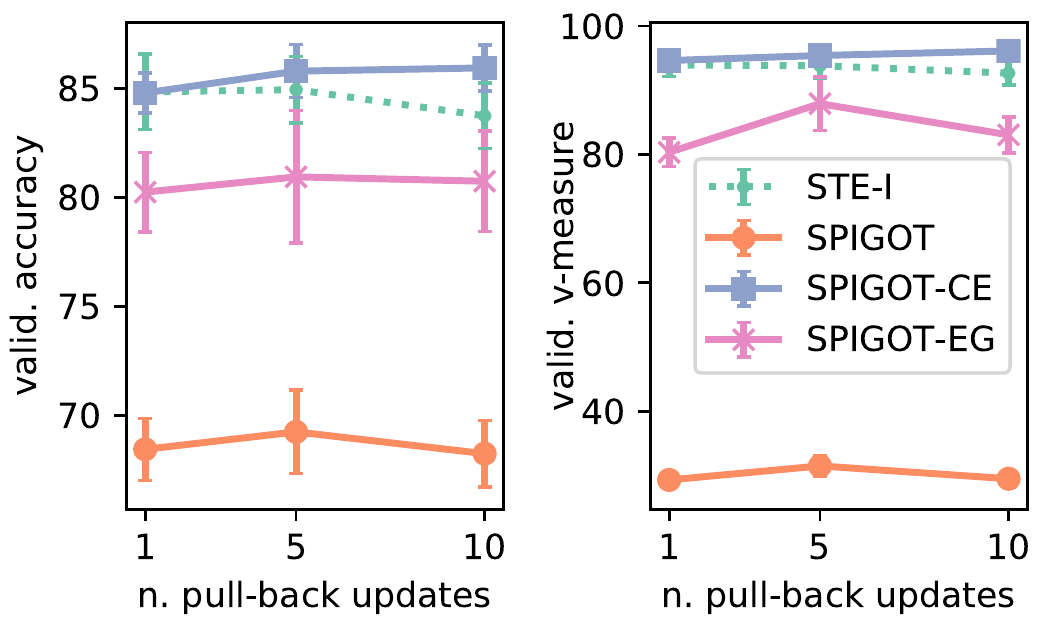}
    \caption{Impact of multiple gradient update steps for the
pulled-back label, on the synthetic example with 10 clusters. For each point,
the best step size $\eta$ is chosen. \label{fig:pullback_steps_comparison}}
\end{figure}

\paragraph{Impact of multiple updates.}
One possible explanation for the failure of SPIGOT is that SPIGOT-CE and SPIGOT-EG
perform more work per iteration, since they use a softmax initial guess and thus require
a second pass through the decoder.
We rule out this possibility in Figure~\ref{fig:pullback_steps_comparison}:
even when tuning the number of updates,
SPIGOT does not substantially improve.
We observe, however, that SPIGOT-CE improves slightly with more updates,
outperforming STE-I. However,
since each update step performs an additional decoder call, this also increases the training time.

\subsection{Structured Latent Variables}\label{subsection:structured_latent_variables}

For learning structured latent variables, we study sentiment
classification on the English language Stanford Sentiment Treebank (SST)
\citep{socher2013recursive}, and Natural Language Inference on the SNLI dataset \citep{snli:emnlp2015}.

\subsubsection{Sentiment Classification}
The model predicts a latent projective arc-factored dependency tree
for the sentence, then uses the tree in predicting the
downstream binary sentiment label.
The model has the following components:

\begin{itemize}
    \item \textbf{Encoder:} Computes a score for every possible dependency arc $i
\rightarrow j$ between words $i$ and $j$.
Each word is represented by its embedding
$h_i$,\footnote{Pretrained GloVe vectors \citep{pennington2014glove}.
}
then processed by an
LSTM, yielding contextual vectors
$\overleftrightarrow{h_i}$.
Then, arc scores are computed as
\begin{equation}
s_{i \rightarrow j} =  v^\top \operatorname{tanh}\big(W^\top [\overleftrightarrow{h_i};
\overleftrightarrow{h_j}] + b\big).
\label{eq:arc_scores}
\end{equation}

\begin{table*}[ht!]
\centering
\small
\begin{tabular}{ l
r@{\color{gray}\small$\pm$}>{\color{gray}\small}l r@{$\color{gray}\small\pm$}>{\color{gray}\small}l
r@{\color{gray}\small$\pm$}>{\color{gray}\small}l r@{$\color{gray}\small\pm$}>{\color{gray}\small}l
}
\toprule
&
\multicolumn{4}{c}{\textbf{SST}} &
\multicolumn{2}{c}{\textbf{SNLI}} \\
\textbf{Model} &
\multicolumn{2}{c}{\textbf{Valid. Acc.}} & 
\multicolumn{2}{c}{\textbf{Test Acc.}} &
\multicolumn{2}{c}{\textbf{Valid. Acc.}} & 
\multicolumn{2}{c}{\textbf{Test Acc.}} \\
\midrule
Baseline  &
83.79	&	0.17	&	83.99	&	0.32 & 
85.54	&	0.14	&	85.09	&	0.21 
\\
\addlinespace[1ex]
\multicolumn{5}{l}{\emph{Relaxed}} \\
Marginals &
84.43	&	0.27	&	83.45	&	0.56  & 
85.60	&	0.11	&	85.01	&	0.11 
\\
SparseMAP &
83.94	&	0.41	&	83.61	&	0.33  & 
85.54	&	0.10	&	\textbf{85.35}	&	0.06 
\\
\addlinespace[1ex]
\multicolumn{5}{l}{\emph{Argmax}} \\
*Perturb-and-MAP &
84.06	&	0.59	&	82.92	&	0.61  & 
84.62	&	0.14	&	83.80	&	0.06 
\\

STE-S &
83.25	&	0.83	&	83.32	&	0.88  & 
82.07	&	0.50	&	81.10	&	0.65 
\\
STE-I &
83.44	&	0.70	&	83.17	&	0.11 & 
81.39	&	0.63	&	81.00	&	0.32 
\\
SPIGOT &
84.51	&	0.80	&	\textbf{84.80}	&	1.10 & 
84.03	&	0.28	&	83.52	&	0.24 
\\
SPIGOT-CE &
82.22	&	0.61	&	83.01	&	0.55 & 
80.22	&	1.02	&	79.20	&	0.68 
\\
SPIGOT-EG &
82.94	&	1.06	&	82.88	&	0.90 & 
85.36	&	0.16	&	84.84	&	0.16 
\\

\bottomrule
\end{tabular}
\caption{
SST and SNLI
average accuracy and standard deviation over three runs, with latent dependency trees.
Baselines are described in Section~\ref{subsection:structured_latent_variables}.
We mark stochastic methods marked with *.
}
\label{table:results_structured}
\end{table*}

\item \textbf{Latent parser:} We use the arc scores vector $s$ to get a parse
$\hat{z} = \rho(s)$ for the sentence, where $\rho(s)$ is the {\tt argmax}, or
combination of trees, such as {\tt Marg} or {\tt SparseMAP}.

\item \textbf{Decoder:}
Following \citet{spigot}, 
we concatenate
each $\overleftrightarrow{h_i}$ with its predicted head $\overleftrightarrow{h}_{\text{head}(i)}$.%
~For relaxed methods,  we average all possible heads,
weighted by the corresponding marginal: $\overleftrightarrow{h}_{\text{head}(i)} \defeq \sum_j \mu_{i \rightarrow j} \overleftrightarrow{h_j}$.
The concatenation is passed through an affine layer, a {\tt ReLU}
activation, an attention mechanism,
and the result is fed into
a linear output layer.
\end{itemize}

\noindent For marginal inference, we use
{\tt pytorch-struct} \citep{alex2020torchstruct}.
For the SparseMAP projection, we use the active set algorithm \citep{sparsemap}.
The baseline we compare our models against is a BiLSTM, followed by feeding the sum of all hidden states to a two-layer ReLU-MLP.
\paragraph{Results.}
The results from the experiments with the different methods are shown in Table \ref{table:results_structured}.
As in the unstructured case, the relaxed models lead to strong downstream classifiers. 
Unlike the unstructured case, SPIGOT is a top performer here.
The effect of tuning the number of gradient update steps is not as big as in the unstructured case and did not lead to significant improvement.
This can be explained by a ``moving target'' intuition: since the decoder $g_\theta$ is far from optimal, more accurate $\mu$
do not overall help learning.

\subsubsection{Natural Language Inference}
We build on top of the decomposable attention model
\citep[DA;][]{parikh2016decomposable}. Following the setup of \citet{Corro2019-acl}, we induce structure on the premise and the hypothesis.
For computing the score of the arc from word $i$ to $j$, we concatenate the representations of the two words, as in Eq.~\ref{eq:arc_scores}.
In the \emph{decoder}, after the latent parse tree is calculated, we concatenate each word with the average of its heads. We do this separately for the premise and the hypothesis.
As baseline, we use the DA model with no intra-attention.

\paragraph{Results.}

The SNLI results are shown in Table~\ref{table:results_structured}. Here, the
straight-through (argmax) methods are outperformed by
the more stable relaxation-based methods.
This can be attributed to the word-level alignment in the DA model, where soft
dependency relations appear better suited than hard ones.

\section{Conclusions} 

In this work, we provide a novel motivation for
straight-through estimator (STE) and SPIGOT,
based on pulling back the downstream loss.
We derive promising new algorithms, and novel insight into existing ones.
Unstructured controlled experiments suggest that our new algorithms,
which use the cross-entropy loss instead of the perceptron loss,
can be more stable than SPIGOT while accurately disentangling the latent
variable.
Differentiable relaxation models (using softmax and sparsemax) are the easiest
to optimize to high downstream accuracy, but they
fail to
correctly identify the latent clusters.
On structured NLP experiments, relaxations
(SparseMAP and Marginals) tend to overall perform better
and be more stable than straight-through variants in terms of
classification accuracy.
However, the lack of gold-truth latent structures makes it
impossible to assess recovery performance.
We hope that our
insights, including some of our negative results, 
may encourage future research on learning with latent structures.

\section*{Acknowledgments}
The authors are grateful to
Caio Corro
and the reviewers
for their valuable suggestions.
This work is built on the open-source scientific Python stack \citep{python,numpy,nparray,scipy,cython,scikit-learn,pytorch}.
This work was %
supported by the European Research Council (ERC StG DeepSPIN 758969),
by the P2020 project MAIA (contract 045909),
and by the Funda\c{c}\~ao para a Ci\^encia e Tecnologia
through contract UIDB/50008/2020.

\bibliography{emnlp2020}
\bibliographystyle{acl_natbib}

\clearpage

\appendix

\section{Training Details}
\label{appendix:training_details}

We trained all models with AdamW optimizer \citep{kingma2014adam, loshchilov2018decoupled}.
The embeddings for the SST and SNLI experiments are initialized with Glove
embeddings of size 300 \citep{pennington2014glove}, available from \url{https://nlp.stanford.edu/projects/glove/}.
The training details for all experiments are described in Table~\ref{table:reproducibility}.

\paragraph{Computing Infrastructure} Each experiment was run on a single GPU. The setup of the computers we used is as follows:
\begin{itemize}
    \item GPU: Titan Xp - 12GB\\
        CPU: 16 x AMD Ryzen 1950X @ 3.40GHz - 128GB
    \item GPU: RTX 2080 Ti - 12GB \\
        CPU: 12 x AMD Ryzen 2920X @ 3.50GHz - 128GB
\end{itemize}

\begin{table*}[ht!]
\small
\centering
\begin{tabular}{lp{3.5cm}p{3.5cm}p{3.5cm}}
     & \textbf{Synthetic Data} & \textbf{SST} & \textbf{SNLI} \\
\toprule

\multicolumn{4}{l}{\emph{Data}} \\
Where to get it & Generation script included &
    \url{https://nlp.stanford.edu/sentiment/} &
    \url{https://nlp.stanford.edu/projects/snli/} \\

Preprocesing & \S\ref{section:unstructured}; attached code.
& Neutral instances removed. &  \\

\addlinespace[1ex]
\multicolumn{4}{l}{\emph{Dataset size}} \\
Training set & 5000 & 6920 & 570K \\
Validation set & 1000 & 872 & 10K \\
Test set & 1000 & 1821 & 10K \\
Labels & 2 & 2 & 3 \\

\addlinespace[1ex]
\multicolumn{4}{l}{\emph{Fixed hyperparameters}} \\
Hidden size & 100 & 100 & 200 \\
Dropout & 0 & 0 & .2 \\
Batch size & one batch & 32  & 64 \\
Number of epochs & 10K & 40 & 40 \\

\addlinespace[1ex]
\multicolumn{4}{l}{\emph{Optimized hyperparameters (maximizing validation accuracy)}} \\
Learning rate ($\times 10^{-3})$ &
$\{ .1, 1, 2 \}$
& $\{.01, .02, .05, .1, .5, 1, 2\}$
& $\{.01, .1, .3, 1, 3, 10\}$\qquad
(keeping $\eta=1)$ \\
Pullback step size $\eta$ &  $\{ .1, 1, 2 \}$ & $\{ .1, 1, 10 \}$ & $\{.001, .01, .1, 1, 10\}$
\qquad
(for best learning rate)
\\

\addlinespace[1ex]
\multicolumn{4}{l}{\emph{Number of model parameters}} \\
Baseline & 2K & 150K & 340K \\
Model with latent structure & 3K & 180K & 420K \\

\addlinespace[1ex]
\multicolumn{4}{l}{\emph{Runtime (minutes)}} \\
Baseline        & $<1$ / 1000 steps & $<1$ / epoch & 1 / epoch \\
Softmax / Marginals       & 1 & 3 & 4 \\
Sparsemax / SparseMAP       & 1 & 3 & 25 \\
Gumbel Softmax / Perturb-and-MAP & 1 & 5 & 7 \\
STE-Softmax / STE-Marginals  & 1 & 4 & 6 \\
STE-Identity    & 1 & 2 & 5\\
SPIGOT          & 1 & 3 & 15\\
SPIGOT-CE       & 2 & 4 & 30 \\
SPIGOT-EG       & 2 & 5 & 7 \\

\addlinespace[1ex]
\multicolumn{4}{l}{\emph{Best learning rate (and pullback step size, where applicable)}} \\
Baseline                         & .001 & .00002& .0001 \\
Softmax / Marginals              & .002 & .0001 & .0001 \\
Sparsemax / SparseMAP            & .001 & .00005& .0003 \\
Gumbel Softmax / Perturb-and-MAP & .002 & .00005& .0001 \\
STE-Softmax / STE-Marginals      & .002 & .00005& .0003 \\
STE-Identity    & .001 & .0001 & .0001 \\
SPIGOT          & .002 (.1) & .0001 (.1) & .0003 (1) \\
SPIGOT-CE       & .001 (.1) & .00005 (.1) & .0001 (.1) \\
SPIGOT-EG       & .001 (.1) & .00005 (.1) & .0001 (.001) \\

\bottomrule
\end{tabular}
\caption{Training details and other reproducibility information.
\label{table:reproducibility}
}

\end{table*}

\section{Examples of Latent Trees}
\begin{figure*}
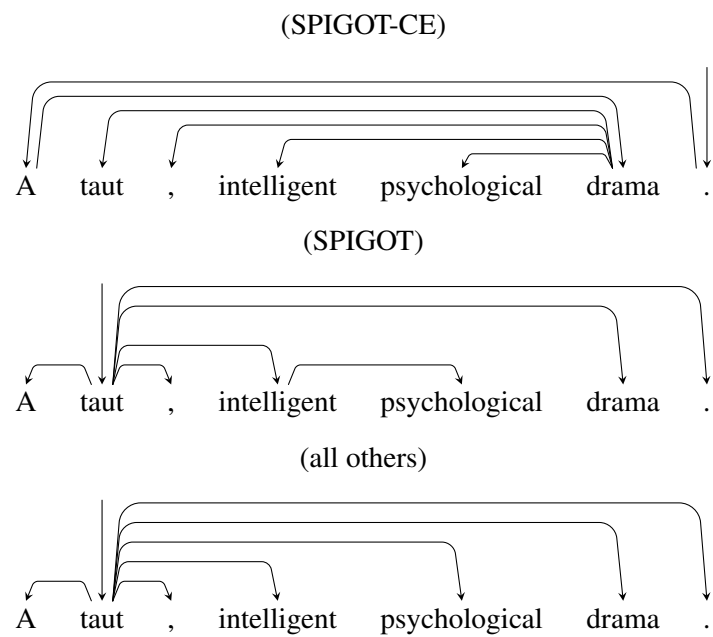

\centering
(SPIGOT-CE)

\begin{dependency}[hide label,edge unit distance=1.1ex]
\begin{deptext}[column sep=0.4cm]
A \&
taut \&
, \&
intelligent \&
psychological \&
drama \&
. \\
\end{deptext}
\depedge{7}{1}{1.0}
\depedge{1}{6}{1.0}
\depedge{6}{2}{1.0}
\depedge{6}{3}{1.0}
\depedge{6}{4}{1.0}
\depedge{6}{5}{1.0}
\deproot[edge unit distance=2.5ex]{7}{}
\end{dependency}


(SPIGOT)

\begin{dependency}[hide label,edge unit distance=1.5ex]
\begin{deptext}[column sep=0.4cm]
A \&
taut \&
, \&
intelligent \&
psychological \&
drama \&
. \\
\end{deptext}
\depedge{2}{1}{1.0}
\depedge{2}{3}{1.0}
\depedge{2}{4}{1.0}
\depedge{4}{5}{1.0}
\depedge{2}{6}{1.0}
\depedge{2}{7}{1.0}
\deproot[edge unit distance=2.5ex]{2}{}
\end{dependency}


(all others)

\begin{dependency}[hide label,edge unit distance=1.5ex]
\begin{deptext}[column sep=0.4cm]
A \&
taut \&
, \&
intelligent \&
psychological \&
drama \&
. \\
\end{deptext}
\depedge{2}{1}{1.0}
\depedge{2}{3}{1.0}
\depedge{2}{4}{1.0}
\depedge{2}{5}{1.0}
\depedge{2}{6}{1.0}
\depedge{2}{7}{1.0}
\deproot[edge unit distance=2.5ex]{2}{}
\end{dependency}

\caption{Example of trees produced by different models for the sentence ``A
taut, intelligent psychological drama.'' The majority of the models produce
mostly flat trees.
In contrast, SPIGOT-CE
identifies the adjectives describing the keyword ``drama'' and attaches them
correctly.\label{fig:example_tree_sst} }

\end{figure*}

We performed a manual analysis of the trees output from the different models.
We notice that, on the SST dataset, most latent trees produced by most models are flat.
This agrees with related work \citep{williams2018latent,sparsemapcg}.
The notable exception is SPIGOT-CE, where the average tree depth on the test set is around 5 and trees seem more informative, suggesting benefits of the cross-entropy loss. Figures~\ref{fig:example_tree_sst_1}, ~\ref{fig:example_tree_sst_2}, ~\ref{fig:example_tree_sst} show examples of the trees produced from different models.

\begin{figure*}
\centering

(SPIGOT-CE)

\begin{dependency}[hide label,edge unit distance=1.1ex]
\begin{deptext}[column sep=0.4cm]
An \&
intelligent \&
, \&
moving \&
and \&
invigorating \&
film \&
. \\
\end{deptext}
\depedge{1}{7}{1.0}
\depedge{2}{3}{1.0}
\depedge{7}{2}{1.0}
\depedge{7}{4}{1.0}
\depedge{7}{5}{1.0}
\depedge{7}{6}{1.0}
\depedge{8}{1}{1.0}
\deproot[edge unit distance=2.5ex]{8}{}
\end{dependency}


(SPIGOT)

\begin{dependency}[hide label,edge unit distance=1.1ex]
\begin{deptext}[column sep=0.4cm]
An \&
intelligent \&
, \&
moving \&
and \&
invigorating \&
film \&
. \\
\end{deptext}
\depedge{1}{2}{1.0}
\depedge{2}{5}{1.0}
\depedge{2}{6}{1.0}
\depedge{2}{7}{1.0}
\depedge{5}{3}{1.0}
\depedge{5}{4}{1.0}
\depedge{8}{1}{1.0}
\deproot[edge unit distance=2.5ex]{8}{}
\end{dependency}


(SPIGOT-EG)

\begin{dependency}[hide label,edge unit distance=1.1ex]
\begin{deptext}[column sep=0.4cm]
An \&
intelligent \&
, \&
moving \&
and \&
invigorating \&
film \&
. \\
\end{deptext}
\depedge{1}{2}{1.0}
\depedge{1}{3}{1.0}
\depedge{1}{4}{1.0}
\depedge{1}{5}{1.0}
\depedge{1}{6}{1.0}
\depedge{1}{7}{1.0}
\depedge{1}{8}{1.0}
\deproot[edge unit distance=2.5ex]{1}{}
\end{dependency}


(STE-I, Marginals, SparseMAP):

\begin{dependency}[hide label,edge unit distance=1.1ex]
\begin{deptext}[column sep=0.4cm]
An \&
intelligent \&
, \&
moving \&
and \&
invigorating \&
film \&
. \\
\end{deptext}
\depedge{6}{1}{1.0}
\depedge{6}{2}{1.0}
\depedge{6}{3}{1.0}
\depedge{6}{4}{1.0}
\depedge{6}{5}{1.0}
\depedge{6}{7}{1.0}
\depedge{6}{8}{1.0}
\deproot[edge unit distance=2.5ex]{6}{}
\end{dependency}

\caption{Example of trees.\label{fig:example_tree_sst_1}}

\end{figure*}

\begin{figure*}
\centering
(SPIGOT-CE)

\begin{dependency}[hide label,edge unit distance=1.1ex]
\begin{deptext}[column sep=0.4cm]
A \&
fascinating \&
and \&
fun \&
film \&
. \\
\end{deptext}
\depedge{1}{5}{1.0}
\depedge{5}{2}{1.0}
\depedge{5}{3}{1.0}
\depedge{5}{4}{1.0}
\depedge{6}{1}{1.0}
\deproot[edge unit distance=2.5ex]{6}{}
\end{dependency}


(SPIGOT)

\begin{dependency}[hide label,edge unit distance=1.1ex]
\begin{deptext}[column sep=0.4cm]
A \&
fascinating \&
and \&
fun \&
film \&
. \\
\end{deptext}
\depedge{2}{1}{1.0}
\depedge{3}{2}{1.0}
\depedge{4}{5}{1.0}
\depedge{4}{6}{1.0}
\deproot[edge unit distance=2.5ex]{3}{}
\deproot[edge unit distance=2.5ex]{4}{}
\end{dependency}

(SPIGOT-EG)

\begin{dependency}[hide label,edge unit distance=1.1ex]
\begin{deptext}[column sep=0.4cm]
A \&
fascinating \&
and \&
fun \&
film \&
. \\
\end{deptext}
\depedge{1}{2}{1.0}
\depedge{1}{3}{1.0}
\depedge{1}{4}{1.0}
\depedge{1}{5}{1.0}
\depedge{1}{6}{1.0}
\deproot[edge unit distance=2.5ex]{1}{}
\end{dependency}

(STE-I, Marginals)

\begin{dependency}[hide label,edge unit distance=1.1ex]
\begin{deptext}[column sep=0.4cm]
A \&
fascinating \&
and \&
fun \&
film \&
. \\
\end{deptext}
\depedge{2}{1}{1.0}
\depedge{2}{3}{1.0}
\depedge{2}{4}{1.0}
\depedge{2}{5}{1.0}
\depedge{2}{6}{1.0}
\deproot[edge unit distance=2.5ex]{2}{}
\end{dependency}

(SparseMAP)

\begin{dependency}[hide label,edge unit distance=1.1ex]
\begin{deptext}[column sep=0.4cm]
A \&
fascinating \&
and \&
fun \&
film \&
. \\
\end{deptext}
\depedge{4}{1}{1.0}
\depedge{4}{2}{1.0}
\depedge{4}{3}{1.0}
\depedge{4}{5}{1.0}
\depedge{4}{6}{1.0}
\deproot[edge unit distance=2.5ex]{4}{}
\end{dependency}


\caption{Example of trees.\label{fig:example_tree_sst_2}}

\end{figure*}

\end{document}